\documentclass[10pt,twocolumn,letterpaper]{article}

\usepackage{cvpr}
\usepackage{times}
\usepackage{epsfig}
\usepackage{graphicx}
\usepackage{amsmath}
\usepackage{amssymb}

\usepackage{amsmath}
\usepackage{multirow}
\usepackage{url}
\usepackage{array}

\usepackage[breaklinks=true,bookmarks=false]{hyperref}

\cvprfinalcopy 

\def\cvprPaperID{****} 

\begin{document}

\title{A Simplified Active Calibration algorithm for Focal Length Estimation}

\author{Mehdi Faraji\thanks{Corresponding author.}, Anup Basu\\
Department of Computing Science,\\ University of Alberta, Canada\\
{\tt\small \{faraji, basu\}@ualberta.ca}}

\maketitle

\begin{abstract}
We introduce new linear mathematical formulations to calculate the focal length of a camera in an active platform. Through mathematical derivations, we show that the focal lengths in each direction can be estimated using only one point correspondence that relates images taken before and after a degenerate rotation of the camera. The new formulations will be beneficial in robotic and dynamic surveillance environments when the camera needs to be calibrated while it freely moves and zooms. By establishing a correspondence between only two images taken after slightly panning and tilting the camera and a reference image, our proposed Simplified Calibration Method is able to calculate the focal length of the camera. We extensively evaluate the derived formulations on a simulated camera, 3D scenes and real-world images. Our error analysis over simulated and real images indicates that the proposed Simplified Active Calibration formulation estimates the parameters of a camera with low error rates.
\end{abstract}
\section{Introduction}
\label{sec:intro}
Many 3D computer vision applications require knowledge of the camera parameters to relate the 3D world to the acquired 2D image(s). The process of estimating the camera parameters is called \textit{camera calibration} in which two groups of parameters (intrinsic and extrinsic) are estimated.

In order to calibrate a camera, conventional calibration methods need to acquire some information from the real 3D world using calibration objects such as grids, wands, or LEDs. This imposes a major limitation on the calibration task since the camera can be calibrated only in off-line and controlled environments. To address this issue, Maybank and Faugeras \cite{maybank1992theory,faugeras1992camera} proposed the so-called \textit{self-calibration} approach in which they used the information of matched points in several images taken by the same camera from different views instead of using known 3D points (calibration objects). In their two-step method, they first estimated the epipolar transformation from three pairs of views, and then linked it to the image of an absolute conic using the Kruppa equations \cite{maybank1992theory}. Not long after the seminal work of Maybank and Faugeras, Basu proposed the idea of Active Calibration \cite{basu1993active2,basu1993active} in which he included the concept of active camera motions and eliminated point-to-point correspondences.

The main downside of the Active Calibration strategies (A and B) in \cite{basu1993active,basu1993active2,basu1995active} is that it calculates the camera intrinsics using a component of the projection equation in which a constraint is imposed by the degenerate rotations. For example, after panning the camera, the equation derived from vertical variations observed in the new image plane is unstable. Furthermore, the small angle approximation using $\sin(\theta) = \theta$ and $\cos(\theta) = 1$ decreases the accuracy of strategies when the angle of rotation is not very small. Also, rolling the camera \cite{basu1997active} is impractical (without having a precise mechanical device) because it creates translational offsets in the camera center. In this paper, we propose a Simplified Active Calibration (SAC) formulation in which the equations are closed-form and linear. To overcome the instability caused by using degenerate rotations in Active Calibration, we calculate focal length in each direction separately. In addition, we do not use small angle approximation by replacing $\sin(\theta) = \theta$ and $\cos(\theta) = 1$. Hence, in our formulation we only refer to the elements of the rotation matrix. Moreover, the proposed method is more practical because it does not require a roll rotation of the camera; only pan and tilt rotations, which can be easily acquired using PTZ cameras, are sufficient.

The rest of the paper is organized as follows. In Section \ref{sec::SAC} we present our proposed Simplified Active Calibration formulation. Section \ref{sec::results} reports and analyzes the results of the proposed method on simulated and real scenes. Finally, conclusions are drawn in Section \ref{sec::conclusion}.

\section{Simplified Active Calibration}\label{sec::SAC}
Simplified Active Calibration (SAC) has been inspired by the novel idea of approximating the camera intrinsics using small angle rotations of the camera which was initially proposed in \cite{basu1993active2,basu1993active} and extended in \cite{basu1995active,basu1997active}. Imposing three constraints on the translation of the camera generates a pure rotation motion. In addition, using small angle rotations allows us to ignore some non-linear terms in order to estimate the remaining linear parameters. The estimated intrinsics can then be used as an initial guess in the non-linear refinement processes.

In general, SAC can be used in any platform in which information about the camera motion is provided by the hardware, such as in robotic applications where the rotation of the camera can be extracted from the inertial sensors or in the surveillance control softwares that are able to rotate the PTZ cameras by specific angles. Having access to the rotation of the camera, we propose a 2-step process to estimate the focal length of the camera. In the first step, we present a closed-form solution to calculate an approximation of the focal length in the $v$ direction ($f_v$) using an image taken after a pan rotation of the camera, assuming that $v$ and $u$ represent the two major axes of the image plane. In the second step, we estimate the focal length of the camera in the $u$ direction ($f_u$) using an image taken after tilt rotation of the camera. Therefore, to estimate the two main components (focal length) of the intrinsic matrix, namely $f_v, f_u$, two pairs of images are required, one taken before and after a small pan rotation, and another taken before and after a small tilt rotation.

\subsection{Focal Length in the v Direction}\label{sec::fv}
We assume that the camera is located at the origin of the Cartesian coordinate system and is looking at distance $z=f$ where the principal point is specified.
Every 3D point $\mathbf{X} = [X\;Y\;Z]^T$ in the world that is visible to the camera can be projected onto a specific point $\mathbf{u} = [v\;u\;1]^T $ of the image plane where the coordinates of the principal points are denoted by $[v_0 \; u_0]^T$. With modern cameras it is reasonable to assume that image pixels are square so that the value of the camera skew is zero.

Every point $\mathbf{u}=[v \; u]^T$ in an image seen by a stationary camera (that freely rotates but stays in a fixed location) is transformed to a point $\mathbf{u'}=[v' \; u']^T$ in another image taken after camera rotation. The mathematical relationship between $\mathbf{u}$ and $\mathbf{u'}$ when the camera is panned is denoted by $w\mathbf{u'}= \mathbf{K}\mathbf{R}_y^{T}\mathbf{K}^{-1}\mathbf{u}$ and after expanding the equation, the relationship is thus represented by:

\begin{equation}\label{eq::fvPrjv}
v' = \dfrac{r_{11}(v-v_0)+r_{31}f_v}{r_{13}\dfrac{v-v_0}{f_v}+r_{33}} + v_0
\end{equation}
\begin{equation}\label{eq::fvPrju}
u' = u_0 - \dfrac{u_0-u}{r_{13}\dfrac{v-v_0}{f_v}+r_{33}}
\end{equation}
Where $r_{ij}$ is an element of the rotation matrix around $Y$-axis at row $i$ and column $j$. After simplification of Eq.\ref{eq::fvPrju}:
\begin{equation}\label{eq::fvPrjSimpleU}
\dfrac{v-v_0}{f_v} = \dfrac{\dfrac{u_0-u}{u_0-u'}-r_{33}}{r_{13}}
\end{equation}

Note that after a pure pan rotation, the $u$ coordinates of the new image will not be affected by the transformation. (The reader is referred to \cite{junejo2012optimizing} for a detailed explanation and analysis about this fact.) In other words, image pixels only move horizontally. Thus, the rate of change in the $u$ direction before and after the pan rotation is close to one, viz:
\begin{equation}\label{eq::fvPrjURate}
\dfrac{u_0-u}{u_0-u'} \approx 1
\end{equation}
Substituting Eq.\ref{eq::fvPrjURate} into Eq.\ref{eq::fvPrjSimpleU} and then replacing the equation obtained for the term $\dfrac{v-v_0}{f_v}$ in the Eq.\ref{eq::fvPrjv}, we have:
\begin{equation}\label{eq::fvPrjvSubs}
v' \approx \dfrac{r_{11}(v-v_0)+r_{31}f_v}{r_{13}\dfrac{1-r_{33}}{r_{13}}+r_{33}} + v_0
\end{equation}
The above substitution changes the value of the denominator to 1 and hence simplifies the whole projection equation.
\begin{equation}\label{eq::fvPrjvFvV0}
v'-r_{11}v \approx r_{31}f_v +(1-r_{11}) v_0
\end{equation}

Knowing that the principal point is close to the center of the image $(c_u=h/2,c_v=w/2)$, where $h$ and $w$ represent the image height and width respectively, we replace $v_0$ with $c_v$ in Eq.\ref{eq::fvPrjvFvV0}. Thus, we can derive a suitable linear equation to estimate the focal length in the $x$ direction from an image taken after a pan rotation.
\begin{equation}\label{eq::fvFinal}
f_v \approx \dfrac{v'-r_{11}v- (1-r_{11})c_v}{r_{31}}
\end{equation}
Eq.\ref{eq::fvFinal} needs only one point $v$ in the reference image that corresponds to $v'$ in the transformed image. If there are more point correspondences, we can easily use the average of these points to obtain more robust results.

\subsection{Focal Length in the u Direction}
So far, we could estimate $f_v$ by the information provided from an image taken after a pan rotation. We repeat the same procedure to approximate $f_u$. This time, we need an image taken after a pure tilt rotation of the camera. Thus, the projection equation is characterized by replacing $\mathbf{R}$ with the proper rotation matrix that describes rotation of the camera around $X$-axis. Following the same reasoning as in Section \ref{sec::fv}, a closed-form solution to estimate the focal length of the camera in the $u$ direction is obtained by:
\begin{equation}\label{eq::fuFinal}
f_u \approx \dfrac{r_{22}u - u' + (1-r_{22})c_u}{r_{32}}
\end{equation}
\begin{figure}[t]\centering
	\begin{tabular}{@{}c@{}c@{}}
		\includegraphics[width=0.48\columnwidth]{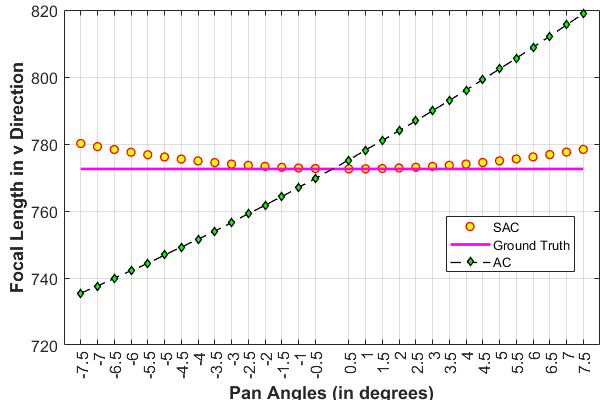}&
		\includegraphics[width=0.48\columnwidth]{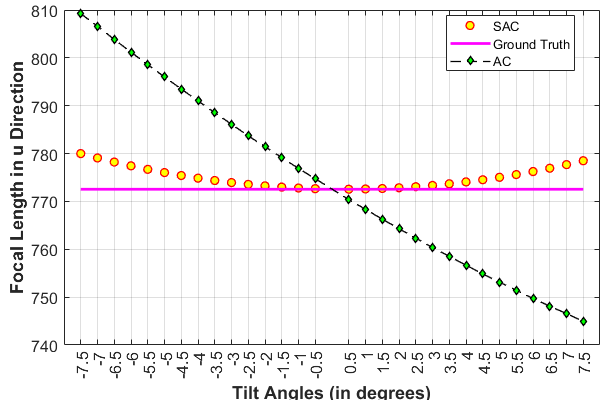}\\
	\end{tabular}	
	\caption{Focal lengths calculated in the $v$ and $u$ directions using Active Calibration Strategy B (AC)\cite{basu1997active} versus SAC for various angles of rotations. In SAC we only use one point correspondence.}
	\label{fig::err_focal}
\end{figure}

\section{Results and Analysis}
\label{sec::results}
Based on our proposed method, focal length in the $v$ and $u$ directions can be estimated using Eq.\ref{eq::fvFinal} and Eq.\ref{eq::fuFinal}, respectively. Only one point correspondence is required to calculate the focal length. Fig.\ref{fig::err_focal} shows the estimated focal lengths using various pan and tilt angles on a 3D synthetic scene of a teapot taken by a simulated camera. It can be seen that when the pan and tilt angles are small, the estimated focal lengths are very close to the ground truth.

In another experiment, we calculate the proposed simplified active calibration formulation on 1000 different runs of 500 randomly generated 3D points for small pan and tilt angles. The mean and standard deviation of the results obtained are shown in Table \ref{tbl::random3D}. As we can see, our proposed active calibration formulation attains results very close to the ground truth. Specifically, the error in focal length estimates is less than 1 pixels.

\subsection{Angular Uncertainty}
Acquiring the rotation angles requires either specific devices such as gyroscopes or a specially designed camera called a PTZ camera. Even using these devices does not guarantee that the extracted rotation angles are noise-free. To simulate the noisy conditions of a real-world application, we contaminated the angles of the above-mentioned teapot sequences with increasing angular errors.

\begin{figure}[t]\centering
	\begin{tabular}{@{}c@{}c}
		\includegraphics[width=0.48\columnwidth]{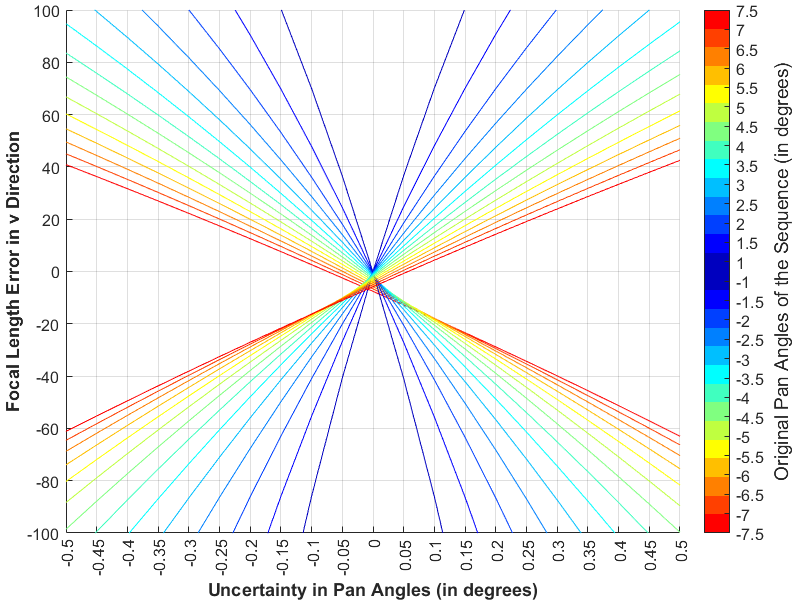}&  \includegraphics[width=0.48\columnwidth]{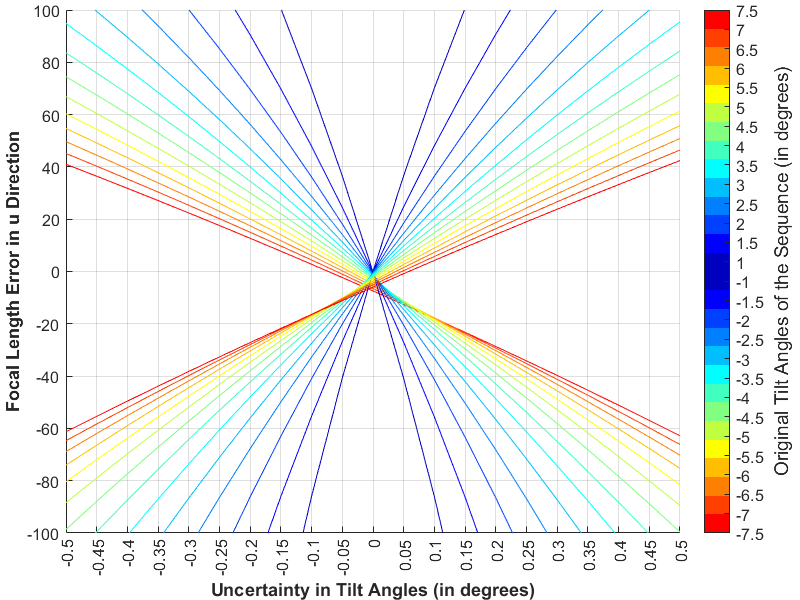}\\
	\end{tabular}
	\caption{The error caused by uncertainty in determining the angle of the camera. Top: The effects of the uncertainty of the camera pan rotation on calculating the focal length in the $v$ direction by SAC. Bottom: The effects of the uncertainty of the camera tilt rotation on calculating the focal length in the $u$ direction by SAC.}
	\label{fig::err_Ang}
\end{figure}

While the point correspondences are kept fixed for all of the pan and tilt rotations, we calculate the focal length (Eq.\ref{eq::fvFinal} and Eq.\ref{eq::fuFinal}) using contaminated pan and tilt angles. The results are shown in Fig.\ref{fig::err_Ang}. Specifically, Fig.\ref{fig::err_Ang}(a) and Fig.\ref{fig::err_Ang}(b) show the error of our proposed formula for estimating the focal length when the pan and tilt angles are not accurate. Every sequence has been colored based on its rotation angle, ranging from blue indicating smaller angles to red for larger angles. For focal length estimation, Fig.\ref{fig::err_Ang}(a) and Fig.\ref{fig::err_Ang}(b) illustrate that the sequences taken with smaller angles have steeper slopes than the sequences acquired with larger rotation angles. This shows that focal lengths are more sensitive to angular noise when the camera is rotated by smaller angles rather than larger angles.

Overall, when the camera is rotated by small angles, the influence of the angular noise on SAC equations is higher. On the other hand, SAC tends to use the benefit of rotating the camera by small angles. Therefore, to avoid magnifying the effect of noise it is important not to rotate the camera by very small angles.

\begin{table}[b]
	\begin{minipage}[b]{1.0\linewidth}\centering
		\caption{Results of the proposed simplified active calibration on 1000 separate 3D random points for various small pan and tilt angles. In the table, GT denotes the Ground Truth, SD represents the Standard Deviation. The error values are in pixels. }
		\label{tbl::random3D}
		
		\begin{tabular}{|ccc|cc|}
			\hline
			Pan & Tilt &   &	$f_v$  &$f_u$  \\
			\hline\hline
			&&GT         &	772.55 &772.55 \\
			\hline
			\multirow{3}{*}{$1^{\circ}$} & \multirow{3}{*}{$-1^{\circ}$} &		Mean                 &	772.61 &772.76  \\
			& & SD   &	0.02   &0.09     \\
			& & Error   &	0.06    &0.21   \\	
			\hline	
			\multirow{3}{*}{$-1.5^{\circ}$} & \multirow{3}{*}{$1.5^{\circ}$} &		Mean                 &	773.02 &772.73 \\
			& & SD   &	0.13   &0.07  \\
			& & Error   &	0.47    &0.19 \\	
			\hline	
		\end{tabular}
	\end{minipage}
\end{table}
\subsection{Point Correspondence Noise}
\begin{figure}[t]\centering
	\begin{tabular}{@{}c@{}c@{}}
		\includegraphics[width=0.48\linewidth]{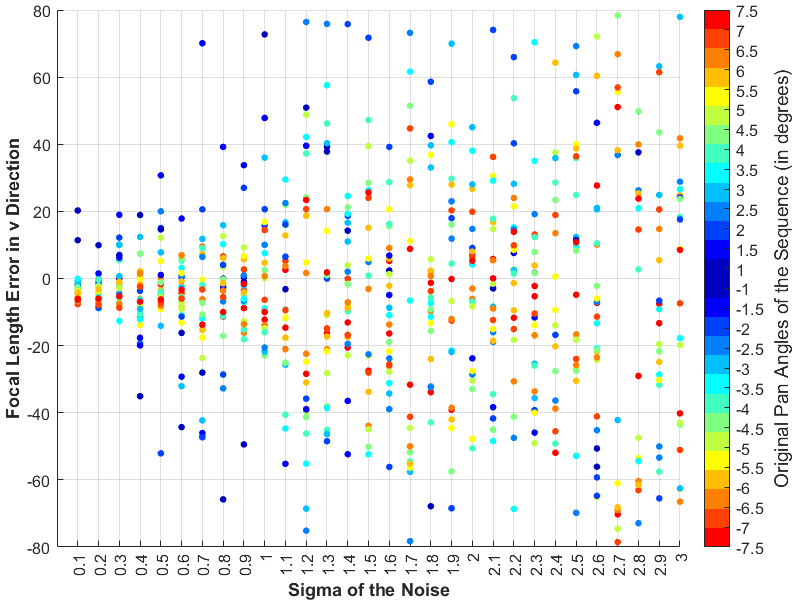} &  \includegraphics[width=0.48\linewidth]{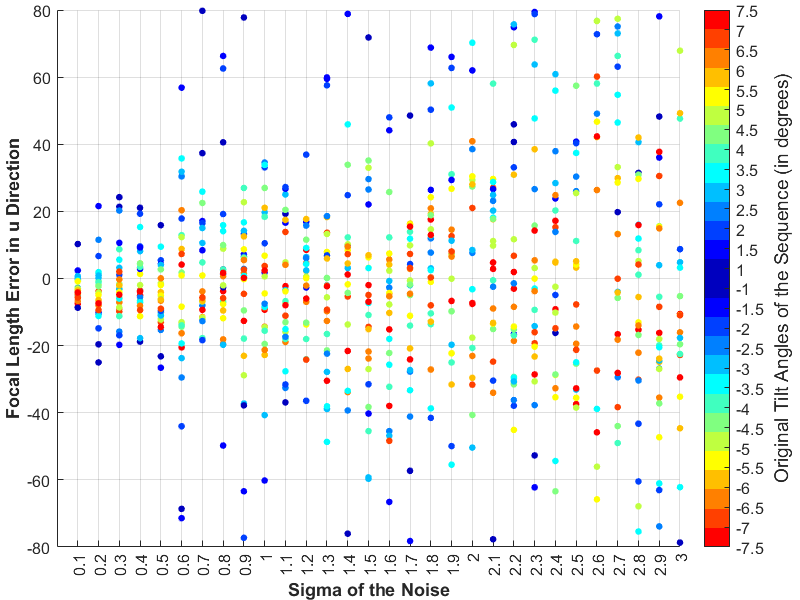}\\
		\textbf{(a)} & \textbf{(b)}\\
	\end{tabular}
	\caption{The error caused by uncertainty in location of points. \textbf{a)} Error of the estimated focal length in the $v$ direction using SAC when the location of the teapot points are disturbed by different values of $\sigma_{pixel}$. \textbf{b)} Error of the estimated focal length in the $u$ direction using SAC under the same conditions as in (a).}
	\label{fig::err_PC}
\end{figure}

Another type of noise that affects the SAC equations is the noise in the location of features used for matching. To simulate such conditions, we assume that the location of every teapot point is disturbed by a Gaussian noise with zero mean and variance $\sigma_{pixel}$. Then, we calibrate the camera using SAC for all $\sigma_{pixel}$ in the range of $0$ to $3$. The intrinsic parameters obtained are illustrated in Fig.\ref{fig::err_PC}.

Fig.\ref{fig::err_PC}(a) and Fig.\ref{fig::err_PC}(b) illustrate the influence of pixel noise on the estimation of focal length (Eq.\ref{eq::fvFinal} and Eq.\ref{eq::fuFinal}). Colors are distributed based on the rotation angles of the camera and, hence, the distribution of the colors reveals how noise affects the SAC equations. In fact, the high concentration of red, yellow, and orange points around the zero error line in Fig.\ref{fig::err_PC}(a) to (b) reveals that when the angle of the camera rotation is not very small, SAC achieves low-error estimates for focal lengths. This corroborates the claim that very small camera rotations can cause results from the SAC formulations to have high error.


\subsection{Real Images}
We studied the proposed SAC formulations on real images as well. We used the Canon VC-C50i PTZ camera that is able to freely rotate around $Y$-axis (pan) and $X$-axis (tilt). The camera can be controlled by a host computer using a standard RS-232 serial communication. Therefore, the required pan and tilt rotation angles can be set in a specific packet and then be written into the camera serial buffer to rotate the camera based on the assigned rotation angles.

\begin{figure}[t]\centering
	\begin{tabular}{@{}c@{}c@{}}
		\includegraphics[width=0.47\linewidth]{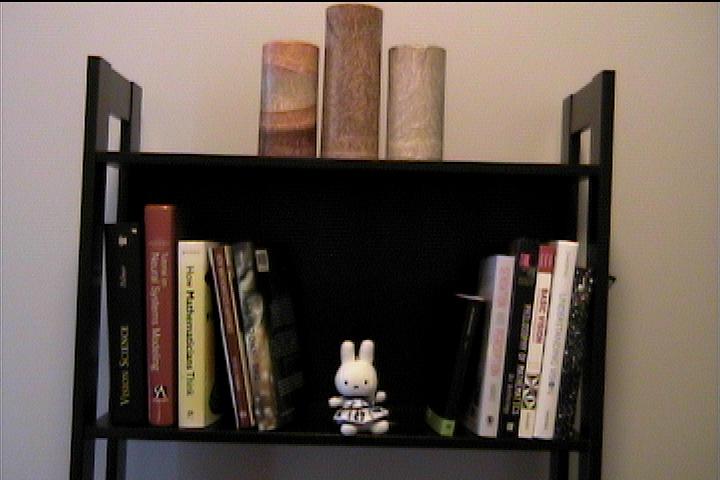}&
		\includegraphics[width=0.47\linewidth]{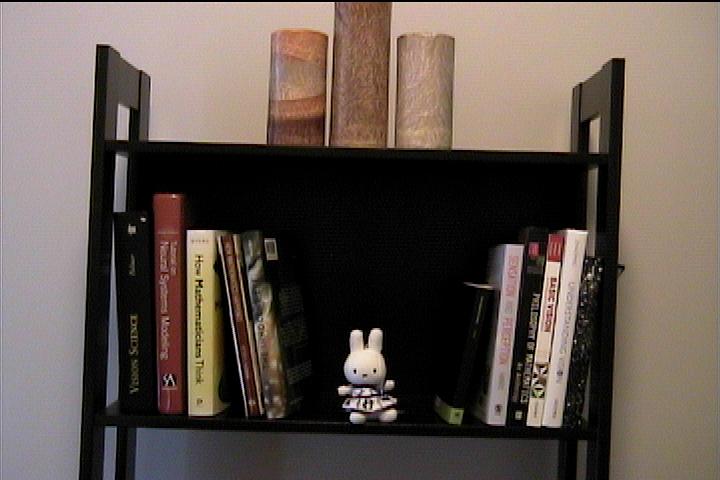} \\
		\textbf{(a)} & \textbf{(b)}\\
	\end{tabular}
	\caption{A sequence of real images taken for SAC. \textbf{a)} Image taken after panning the camera by $0.5625^\circ$. \textbf{b)} Image taken after tilting the camera by $-0.675^\circ$.}
	\label{fig::real_images}
\end{figure}

Using the above-mentioned procedure, we captured four sequences of images for evaluating the proposed SAC formulations. Fig.\ref{fig::real_images} shows a sequence of our bookshelf scene. All sequences were taken using a fixed zoom. While keeping the zoom of the camera unchanged, another 30 images were acquired from various viewpoints of a checkerboard pattern. The ground truth of intrinsic parameters were found by applying the method of Zhang \cite{zhang1999flexible} on the checkerboard images.

The performance of SAC formulations on the four sequences of real images is reported in Table \ref{tbl::RealImages}. For every sequence, we only used the images in the sequence. For example, to calculate the focal length in the $v$ direction of Sequence 1, we found the point correspondence between a reference image and the image taken after the pan rotation of the camera (Fig.\ref{fig::real_images}(a)). Then, we used only one of the matched points that is closer to the center of the image. Although we did not include the lens distortion parameter into the SAC formulation (because it creates non-linear equations), we decrease the inaccuracy of the focal length estimates by using a matched point that is closer to the center of the image and, thus, is less affected by the lens distortion. A similar procedure was adopted with the image taken after a tilt rotation of the camera (Fig.\ref{fig::real_images}(b)) for calculating the focal length in the $u$ direction of Sequence 1.

The errors reported by applying SAC on four different sequences of real images in Table \ref{tbl::RealImages} show that despite the presence of various types of noise, such as angular uncertainties, point correspondence noise and lens distortion, focal lengths estimated by SAC are close to the results of the method of Zhang \cite{zhang1999flexible}, except when the angles of rotations are very small ($< 1^\circ $).
\section{Conclusion}\label{sec::conclusion}
Inspired by the idea of calibrating a camera through active movements of the camera, in this paper we presented a Simplified Active Calibration formulation. Our study provides closed-form and linear equations to estimate the parameters of the camera using two image pairs taken before and after panning and tilting the camera.

A basic assumption about the rotation of a fixed camera was made; i.e., to solve the proposed equations, knowing the rotation angles of the camera is necessary. The proposed formulation can be used in practical applications such as surveillance, because in PTZ and mobile phone cameras accessing the camera motion information is straightforward.
\begin{table}[b]
	\caption{Results of the proposed simplified active calibration on four sequences of real images. All angles are in degrees. $\delta_{f_v}$,$\delta_{f_u}$ are the percentage errors from the corresponding ground truth acquired by the method of Zhang \cite{zhang1999flexible}. }
	\label{tbl::RealImages}
	\begin{minipage}[b]{1.0\linewidth}\centering	
		\begin{tabular}{|ccccccc|}
			\hline
			\#	&	  Pan   &    Tilt    &   $f_v$   &  $f_u$    &   $\delta_{f_v}$  &   $\delta_{f_u}$ \\
			\hline
			1 &		 0.5625${^\circ}$ &    -0.675${^\circ}$  &   880.42  &  -999.07 &  15.3  &  5.33  \\
			2 &		   -1.8${^\circ}$ &     2.025${^\circ}$  &   1052.1  &  -966.35 &  1.18  &  1.88  \\
			3 &		-4.6125${^\circ}$ &   -4.1625${^\circ}$  &   1067.9  &     -970 &  2.70  &  2.26  \\			
			4 &		-7.9875${^\circ}$ &    -7.425${^\circ}$  &   1069.6  &  -986.58 &  2.87  &  4.01  \\
			\hline
		\end{tabular}
	\end{minipage}
\end{table}

The proposed closed-form formulations for estimating the focal lengths can be solved with only one point correspondence. Finding the correspondence point is straightforward. Due to the recent developments in feature extractors, one may use \cite{faraji2015erel,faraji2015extremal} to extract repeatable regions from a pair of images. This is especially useful for applications that prefer no point correspondences; where instead of the reference and transfered points in Eq.\ref{eq::fuFinal} and Eq.\ref{eq::fvFinal}, the average of the edge points or the centroid of the regions can be used.

The results of solving our proposed formulations on randomly simulated 3D scenes indicated a very low error rate in estimating the focal lengths. We evaluated our proposed SAC formulation for two different noise conditions, namely angular and pixel noise. The simulated results showed that if the angle of rotation is not very small, the SAC formulation can robustly estimate the focal lengths. This conclusion was later verified in our experiment with real images. Our future work will focus on deriving linear equations for calculating the location of the principal point and also including lens distortion parameters into the Simplified Active Calibration equations.

{\small
\bibliographystyle{ieee}
\bibliography{refs}
}

\end{document}